\documentclass[10 pt, conference]{ieeeconf}  %

\IEEEoverridecommandlockouts                              %

\usepackage{cite}
\usepackage{times}
\usepackage{mathtools}
\usepackage{multicol}
\usepackage[bookmarks=true]{hyperref}
\usepackage{cleveref}
\usepackage{graphicx}
\usepackage{array,multirow,booktabs} %
\graphicspath{{figures/}}
\usepackage{caption}
\usepackage{subcaption}
\usepackage{amsmath,amssymb} %
\usepackage{pifont}  %

\usepackage[dvipsnames]{xcolor}
\definecolor{SigmaColor}{rgb}{0.98,0.45,0.0}

\title{\LARGE \bf
Learning to Transfer Human Hand Skills for Robot Manipulations
}

\author{Sungjae Park$^{*,1,2}$, Seungho Lee$^{*,2}$, Mingi Choi$^{*,2}$, Jiye Lee$^{2}$, Jeonghwan Kim$^{2}$, Jisoo Kim$^{2}$,  Hanbyul Joo$^{2}$%
\thanks{*equal contribution}%
\thanks{$^{1}$,
        Carnegie Mellon University, $^{2}$ Seoul National University}%
}

\begin{document}

\maketitle
\thispagestyle{empty}
\pagestyle{empty}

\begin{abstract}
We present a method for teaching dexterous manipulation tasks to robots from human hand motion demonstrations. Unlike existing approaches that solely rely on kinematics information without taking into account the plausibility of robot and object interaction, our method directly infers plausible robot manipulation actions from human motion demonstrations.
To address the embodiment gap between the human hand and the robot system, our approach learns a joint motion manifold that maps human hand movements, robot hand actions, and object movements in 3D, enabling us to infer one motion component from others. 
Our key idea is the generation of pseudo-supervision triplets, which pair human, object, and robot motion trajectories synthetically. Through real-world experiments with robot hand manipulation, we demonstrate that our data-driven retargeting method significantly outperforms conventional retargeting techniques, effectively bridging the embodiment gap between human and robotic hands. 

\noindent\textcolor{purple}{\url{https://rureadyo.github.io/MocapRobot/}
}\end{abstract}

\vspace{-5pt}
\section{Introduction}
\label{sec:introduction}
\vspace{-3pt}

Recent advances in imitation learning (IL) via expert demonstrations have significantly improved dexterous manipulation with multi-fingered robotic hands~\cite{handa2020dexpilot, sivakumar2022robotic, qin2023anyteleop, arunachalam2023dexterous, qin2022dexmv}. 
These demonstrations typically come from robotic teleoperation, where a human teleoperates a robot hand using motion capture gloves or a vision-based hand estimation module. While these methods provide physical plausibility of the collected data, collecting demonstrations via teleoperation is often costly, time-consuming, and requires sophisticated skills to operate the robot hardware, which can vary in performance across operators due to the structural limitations of the system. In contrast, demonstrations via human motion capture offer a more natural and convenient alternative. Recent advances in vision-based methods promise more accessible solutions~\cite{pavlakos2024reconstructing, rong2021frankmocap, lee2024mocapevery}, allowing users to perform tasks casually and potentially generating a larger volume of data. Motion capture data also provide rich hand information about how to manipulate the object, which is crucial for dexterous manipulation.

However, transferring human motion demonstrations to robots is not straightforward due to the embodiment gap between human and robotic hands. Differences in skeletal structure, hand size, and forces that can be applied present significant challenges in directly applying imitation learning strategies. %
Consequently,  traditional retargeting methods that transfer human demonstrations to robotic hands via, for example, kinematics-based~\cite{handa2020dexpilot, qin2022dexmv, wang2024dexcap} alignment often yield suboptimal results, leading to task failures. An ideal retargeting should be \textbf{generalizable} to different hand/object motions to benefit from the scalability of human motion demonstrations, and also output \textbf{physically plausible actions } for the robot hand under such embodiment gap.

In this paper, we propose a novel approach to inferring plausible robot hand actions from human motion demonstrations, through a learning-based retargeting method. Specifically, our aim is to find the mapping between the robot hand actions and human hand motion to achieve the same target object motion. To achieve this, we formulate the problem within a supervised learning framework,  learning a joint manifold space among human hand motion, robot action, and 3D object movements in a data-driven manner. The primary challenge in learning such mapping is the lack of available paired datasets for common desired actions. To address this, we introduce a method to synthetically generate paired human-robot grasping data by combining separately captured human motion capture demonstrations and teleoperation data on the same target object. To this end, our method can infer the effective robot action trajectory from human manipulation demonstration by finding an optimal latent code of the manifold space toward the provided human and object motion trajectories. Through extensive evaluations, we demonstrate the effectiveness of our approach in solving complex dexterous manipulation tasks.

\begin{figure}[t]
\centering
\includegraphics[width=0.95\columnwidth, trim={0cm 1.3cm 0cm 0.2cm}]{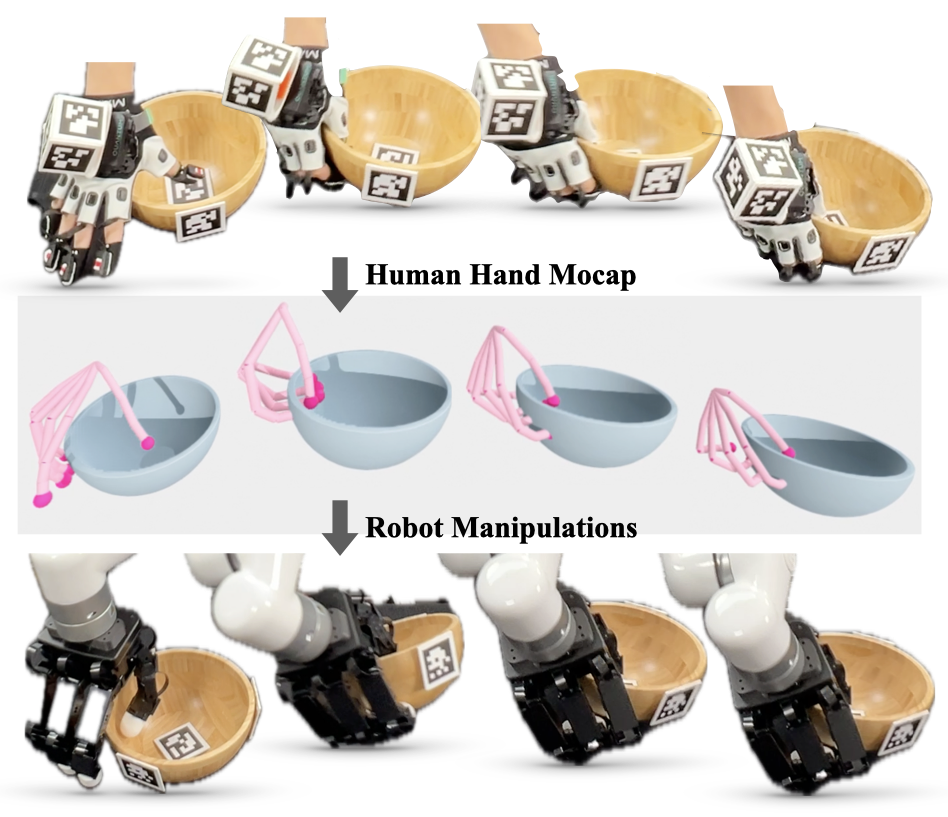}
    \caption{Our model learns a \textbf{human-to-robot retargeting model} using an \textbf{unpaired} (i.e., object may move differently) human mocap and robot teleoperation dataset.}
\label{fig:teaser}
\vspace{-22pt}
\end{figure}

\vspace{-5pt}

\section{Related Work}
\label{sec:related_work}

\vspace{-5pt}

\noindent \textbf{Robotic Teleoperation.} In recent years, robotic teleoperation has emerged as a common data source for robotics. Several work proposed a teleoperation system across different robotic platforms, ranging from a dexterous robot hand~\cite{handa2020dexpilot, sivakumar2022robotic, qin2023anyteleop, arunachalam2023dexterous}, mobile robot~\cite{dass2024telemoma}, bimanual robot~\cite{cheng2024open, yang2024ace, ding2024bunny, shawbimanual}, and a humanoid robot~\cite{he2024omnih2o,he2024learning, fu2024humanplus}. Such systems often consist of a perception module which estimates human teleoperator's motion and a retargeting algorithm that kinematically maps human motion to robot actions. While teleoperation data contain rich information for training robotic policy via imitation learning, it has two major disadvantages: a necessity of hardware system (i.e., robot and teleoperation device such as VR devices~\cite{arunachalam2023holo, cheng2024open},exoskeleoton~\cite{yang2024ace} and additional linkage system~\cite{wu2023gello}) and an embodiment gap between human teleoperator and the robot. Among the two, the latter often receives less attention while significant. Specifically, as the human teleoperateor indirectly interacts with the environment through the system, the teleoperator should fill the embodiment gap via visual feedback(i.e., see whether the robot is interacting with the environment in a desired way) during data collection. If the robot is not acting as intended, the teleoperator should implicitly adapt the teleoperation strategy over time, making teleoperation less scalable when it comes to complex manipulation tasks.  

\noindent \textbf{Retargeting Human Hand to Robot Hand.} Retargeting human's motion or interaction with the environment to that of robot has been a challenging research topic\cite{li2023ace,sermanet2018time, kim2022humanconquad, li2023crossloco, wu2024one
}. The most relevant to ours is retargeting human hand motion to robot hand. ~\cite{handa2020dexpilot, qin2022dexmv, qin2023anyteleop, wang2024dexcap} use an optimization algorithm via kinematic loss, which aims to match certain human joints' position(e.g. fingertips) with that of corresponding robot joints. Temporal consistency loss and self-collision loss are also optionally taken into account.~\cite{sivakumar2022robotic} learn an energy model instead of performing optimization with a similar learning objective. While such methods work well for teleoperation, they may fail when the aim is to reproduce the interaction between human hand and the object, as matching joint positions does not necessarily result in same interactions. To consider the interaction between human hand and the object, ~\cite{du2022multi} additionally considers contact heatmap given human hand and object mesh, so that the retargeted robot hand can grasp the same object similar to human, while being limited to grasping. ~\cite{lakshmipathy2022contact, lakshmipathy2024kinematic} also consider contact regions between human hand and object and try to match it within target robot and same object, but requires human expert labels, such as the center of contact region within the target robot, corresponding points between human hand and robot, etc. Although matching contact regions may result in a more physically plausible robot motion along the object, it does not guarantee the contact region from human mocap to be perfectly matched, and even so, the embodiment gap between human hand and robot hand may result different outcomes to the target object. In short, prior work implicitly assumes that matching kinematical constraints or contact regions would result in same robot-object interaction, which may not always hold. Our work differs from previous work that we directly aim to reproduce the interaction without such assumptions. Specifically, we learn a retargeting model which outputs robot actions that can achieve the same object trajectory given from human mocap data when executed.

\noindent \textbf{Dexterous Manipulation.} Several work have leveraged human mocap data to learn a dexterous robot hand policy, either using it as a target task demonstration\cite{qin2022dexmv,wang2024dexcap}, or as a large dataset to extract general prior\cite{shaw2023videodex,srirama2024hrp}.   
 ~\cite{qin2022dexmv, wang2024dexcap} use human mocap data of a target task as a demonstration for training robot policy, while requiring finetuning with reinforcement-learning or teleoperation data.
~\cite{shaw2023videodex} use a perception module to extract human hand and finger poses from Internet videos \cite{damen2018scaling}, and learn a prior model which outputs plausible robot hand motions given visual input. \cite{srirama2024hrp} additionally extracts 2D contact locations and active object bounding box labels on top of human poses from human video dataset, which is used to train a visual encoder. The most similar setup to ours is~\cite{liu2024quasisim}, where the aim is to find a sequence of robot actions given human mocap data through optimizing a parametrized quasi-physical simulator. However, its main focus is within simulation, lacking real world evaluations. Additionally, as it aims to find a sequence of robot actions given a single sequence of human and object motion, the retargeting process can be computationally expensive when given with a set of different motions of a single object being manipulated by the human hand. Our work directly aims to perform dexterous manipulation in the real world, and shows generalization capabilities to unseen motion of the given object and human hand.

\begin{figure*}[ht]
\centering
\includegraphics[width=1.0\textwidth]{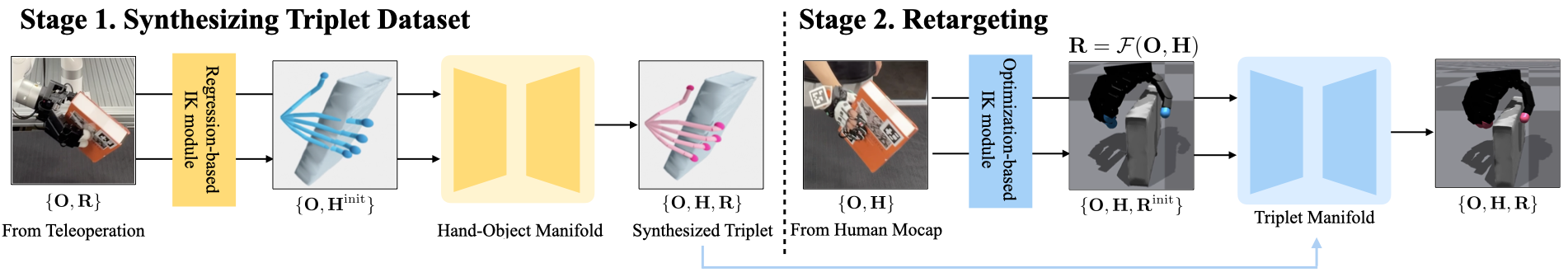}
\vspace{-15px}
    \caption{\textbf{Overview of the Proposed Framework.} We first synthesize the paired triplet dataset consisting of robot action and human motion achieving the same object trajectory, followed by learning a retargeting module. The retargeting model is evaluated in real world, and we use IsaacGym simulator for visualization only.}
\label{fig:method}
\vspace{-15px}
\end{figure*}

\vspace{-6pt}
\section{Method}
\label{sec:method}
\vspace{-5pt}
We propose a learning-based framework that transforms human hand manipulation demonstrations into a sequence of robotic actions, enabling the robot hand to accurately imitate the same manipulation task. Specifically, our method takes as input a target object trajectory $\mathbf{O} \in \mathbb{R}^{o \times t}$ and a human hand demonstration $\mathbf{H} \in \mathbb{R}^{h \times t}$, and it outputs the corresponding plausible robot action trajectory $\mathbf{R} \in \mathbb{H}^{r \times t}$:
\begin{equation}
    \mathbf{R} = \mathcal{F}(\mathbf{O}, \mathbf{H}) 
\label{eq:our_retarget_framework}
\end{equation}
, where $t$ represents the number of timesteps, $o = 9\,$ (3 for position, 6 for rotation),
$h = 81\,$ (3 for human wrist position, 6 for human wrist rotation, and $72 = 24\times3$ for each finger joint's 3D position w.r.t. human wrist frame), and $r = 25\,$ (3 for desired robot wrist position, 6 for desired robot wrist rotation, and 16 for desired robot hand joint angles), respectively. 
We use the 6D rotation representation~\cite{zhou2019continuity}.

Our framework $\mathcal{F}$ is built by learning the joint spatio-temporal manifold over $\mathbf{O}$, $\mathbf{H}$, and $\mathbf{R}$ by training a convolutional autoencoder model~\cite{holden2016deep, lee2023lama}:
\begin{equation}
    (\mathbf{O}, \mathbf{H}, \mathbf{R}) \approx  \mathbf{\Psi}_{\text{dec}} \left(\mathbf{\Psi}_{\text{enc}}\left(\mathbf{O}, \mathbf{H}, \mathbf{R}\right) \right),
\label{eq:manifold_eq}
\end{equation}
where $\mathbf{\Psi}_{\text{enc}}$ and $\mathbf{\Psi}_{\text{dec}}$ are the 1-D temporal convolution encoder and decoder applied to the concatenated triplet $\left(\mathbf{O}, \mathbf{H}, \mathbf{R}\right)$. 
The encoded bottleneck layer $\mathbf{\Psi}_{\text{enc}}\left(\mathbf{O}, \mathbf{H}, \mathbf{R}\right) = \mathbf{L}$ represents the manifold latent code modeling the correlations among human hand, robot action, and target object trajectory during manipulation. This learned manifold enables us to estimate missing components through an optimization-based framework. For example, given human hand motion $\mathbf{H}$ and target object trajectory $\mathbf{O}$, we infer the corresponding plausible robot hand trajectory $\mathbf{R}$ by optimizing the latent code $\mathbf{L}$ as follows: 
\begin{equation}
    \mathbf{L}\textsuperscript{*} =  \underset{\mathbf{L}}{\arg\min}{
    \| \mathbf{\Psi}_{\text{dec}}^{\mathbf{O}}(\mathbf{L})  - \mathbf{O}\|_2}, 
\end{equation}
where $\mathbf{\Psi}_{\text{dec}}^{\mathbf{O}}$ is the object trajectory component decoded from the manifold latent code $\mathbf{L}$, which is compared to the desired object trajectory $\mathbf{O}$. Once the optimal $\mathbf{L}\textsuperscript{*}$ is found, the desired robot hand motion $\mathbf{R}$ can be computed as:
\begin{equation}
    \mathbf{R} = \mathbf{\Psi}_{\text{dec}}^{\mathbf{R}}(\mathbf{L}\textsuperscript{*}) 
\label{eq:final_solution_r}
\end{equation}
, by applying the decoder and extracting the robot hand component from the output, denoted as $\mathbf{\Psi}_{\text{dec}}^{\mathbf{R}}$.
Since the latent code optimization Eq.~\ref{eq:manifold_eq} is performed via a gradient decent method, choosing a good initial $\mathbf{L}\textsuperscript{init}$ is important. To achieve this, we first estimate an initial robot hand motion $\mathbf{R}\textsuperscript{init}$ from human motion $\mathbf{H}$ using a conventional Inverse Kinematics (IK)-based optimization, by matching robot fingertip positions to human fingertip positions. We then use this as the input for the encoder to initialize the latent code: $\mathbf{L}\textsuperscript{init} = \mathbf{\Psi}_{\text{enc}}(\mathbf{O}, \mathbf{H}, \mathbf{R}\textsuperscript{init} )$.

To learn the manifold space over $\left(\mathbf{O}, \mathbf{H}, \mathbf{R}\right)$, it is necessary to collect paired data for the supervision, consisting of human and robot hand motions that result in the same object manipulation. However, it is infeasible to obtain such dataset, as both demonstrations cannot be performed simultaneously. Our key insight is to synthesize plausible pseudo-ground-truth pairs, which we describe next.

\subsection{Synthesizing Pseudo-GT Triplet DB.}
To learn the manifold space described in Eq.~\ref{eq:manifold_eq}, we need a dataset containing a set of triplets $\{\mathbf{O}_i, \mathbf{H}_i, \mathbf{R}_i\}_{i=1}^N$. However, collecting such a dataset is impractical. 
Our solution is to synthesize corresponding human hand motion samples $\mathbf{H}_i$ based on a collected robot teleoperation data $\{\mathbf{O}_i, \mathbf{R}_i\}$.

Specifically, given a target object, we collect two separate datasets: via human mocap manipulation demonstrations $\{\mathbf{O}^M_j, \mathbf{H}^M_j\}_{ j = 1}^M$, and via teleoperation $\{\mathbf{O}_i, \mathbf{R}_i\}$. Then, we build a framework $\mathcal{E}$ to synthesize hand motion as follows:
\begin{equation}
    \mathbf{H}_i = \mathcal{E}(\mathbf{O}_i, \mathbf{R}_i)
\end{equation}
, where our framework $\mathcal{E}$ is composed of two modules: (1) a regressor $\Omega$ to estimate an initial human hand motion $\mathbf{H}^{\text{init}}_i = \Omega(\mathbf{R}_i)$, and (2) a manifold-based refinement process $\psi$ to improve the $\mathbf{H}^{\text{init}}_i$ considering the object trajectory $\mathbf{O}_i$, achieved via manifold learning similar to Eq.~\ref{eq:manifold_eq}. We describe each module below. 

\noindent \textbf{Learning The Manifold Space for $\{\mathbf{O}_i, \mathbf{H}_i\}$.} We first learn the joint spatio-temporal manifold space over $\mathbf{O}$ and $\mathbf{H}$, similar to Eq.~\ref{eq:manifold_eq}, by training an a temporal-convolutional autoencoder model~\cite{holden2016deep, lee2023lama} using the human mocap manipulation dataset $\{\mathbf{O}^M_j, \mathbf{H}^M_j\}$:
\begin{equation}
    (\mathbf{O}, \mathbf{H}) \approx  \mathbf{\psi}_{\text{dec}} \left(\mathbf{\psi}_{\text{enc}}\left(\mathbf{O}, \mathbf{H}\right) \right)
\label{eq:manifold_ho_eq}
\end{equation}
, where the bottleneck latent code $\mathbf{l} = \mathbf{\psi}_{\text{enc}} \left(\mathbf{O}, \mathbf{H}\right)$ captures the spatio-temporal correlation between object trajectory $\mathbf{O}$ and the corresponding hand motion $\mathbf{H}$. Once trained, this model can be used to infer the corresponding hand motion $\mathbf{H}_i$, given object trajectory $\mathbf{O}_i$ by finding the optimal latent code $\mathbf{l}_i\textsuperscript{*}$:
\begin{equation}
    \mathbf{l}_i\textsuperscript{*} =  \underset{\mathbf{l}}{\arg\min}{
    \| \mathbf{\psi}_{\text{dec}}^{\mathbf{O}}(\mathbf{l})  - \mathbf{O}_i\|_2}
\label{eq:manifold_ho_loss}
\end{equation}
, where $\mathbf{\psi}_{\text{dec}}^{\mathbf{O}}$ is the object trajectory component decoded from the manifold latent code $\mathbf{l}$.
Once we obtain the optimal $\mathbf{l}_i\textsuperscript{*}$, the desired human hand motion $\mathbf{H}_i$ can be obtained as:
\begin{equation}
    \mathbf{H}_i = \mathbf{\psi}_{\text{dec}}^{\mathbf{H}}(\mathbf{l}_i\textsuperscript{*}).
\end{equation}
As in optimization for Eq.~\ref{eq:manifold_eq}, selecting a good initial latent code $\mathbf{l}\textsuperscript{init}$ is important. Thus, we first estimate the initial hand motion $\mathbf{H}_i\textsuperscript{init}$ from robot action $\mathbf{R}_i$, and apply it to the pre-trained encoder $\mathbf{l}_i\textsuperscript{init} = \mathbf{\psi}_{\text{enc}}(\mathbf{O}, \mathbf{H}_i\textsuperscript{init} )$. However, unlike the previous case where a traditional IK solver is used, here we train a neural regressor to estimate $\mathbf{H}_i\textsuperscript{init}$.

\noindent \textbf{Regressing Hand Motion from Robot Action.}
One way to estimate the initial human hand motion $\mathbf{H}_i\textsuperscript{init}$ from the provided robot hand $\mathbf{R}_i$ (obtained via teleoperation) is through a traditional Inverse Kinematics (IK), aligning corresponding fingertip and joint positions. However, we found this IK optimization from robot hand to human hand unstable unlike the opposite direction, human hand to robot hand, since the human hand typically has a higher degree of freedom with more fingers. 
As a solution, we build a neural regressor $\mathbf{H}_i\textsuperscript{init} = \Omega(\mathbf{R}_i)$, which we train on a dataset of paired human and robot hand motions $\{\mathbf{H}_k, \mathbf{R}_k\textsuperscript{IK}\}$. To build the paired DB, we compute the robot hand $\mathbf{R}_k\textsuperscript{IK}$ from human hand $\mathbf{H}_k$ using traditional IK optimization, minimizing the difference between human fingertip and robot fingertip positions. 
To represent the human hand and robot hand, we include the rotation and position of the wrist, and the position of the finger joint and fingertips (24 positions for the human hand and only 4 fingertips for the robot defined w.r.t. the wrist frame). 
Then, we train the neural network regressor to predict human hand motion from robot action, $\mathbf{H}_i\textsuperscript{init} = \Omega(\mathbf{R}_i)$.
Our regression model $\Omega$ consists of 6 layers of multi-head self-attention\cite{vaswani2017attention} with 8 heads, each with an embedding dimension of 256.
$\Omega$ operates per frame, converting each robot hand configuration to human hand pose.

\noindent \textbf{Hand Motion Refinement.}
While the initial estimate of human hand motion $\mathbf{H}_i\textsuperscript{init}$ produced by $\Omega$ shows a certain level of visual plausibility, the quality is limited due to the limited quality of the supervision derived from traditional IK, and, more importantly, its failure to account for interactions between hand and object trajectories, by taking only robot hand as the input. 
Our manifold-based optimization using the model of Eq~\ref{eq:manifold_ho_eq} significantly enhances the quality of hand motion synthesis, by capturing the relationship between human hand motion and object movement, resulting in the final output $\mathbf{H}_i$.

While motion manifold autoencoder Eq~\ref{eq:manifold_ho_eq} is trained with a fixed window size, it can be applied to arbitrary lengths of $\{O_{i},H_{i}\}$ by applying it at each starting point in a sliding window fashion, and optimizing latent codes at all time window together via Eq~\ref{eq:manifold_ho_loss}, along with enforcing temporal consistency of the overlapped output.

\vspace{-5pt}
\subsection{Hardware System Setup for Data Collection}
\begin{figure}[t]
\centering
\includegraphics[width=0.9\columnwidth,  trim={0cm 0.6cm 0cm 0cm}]{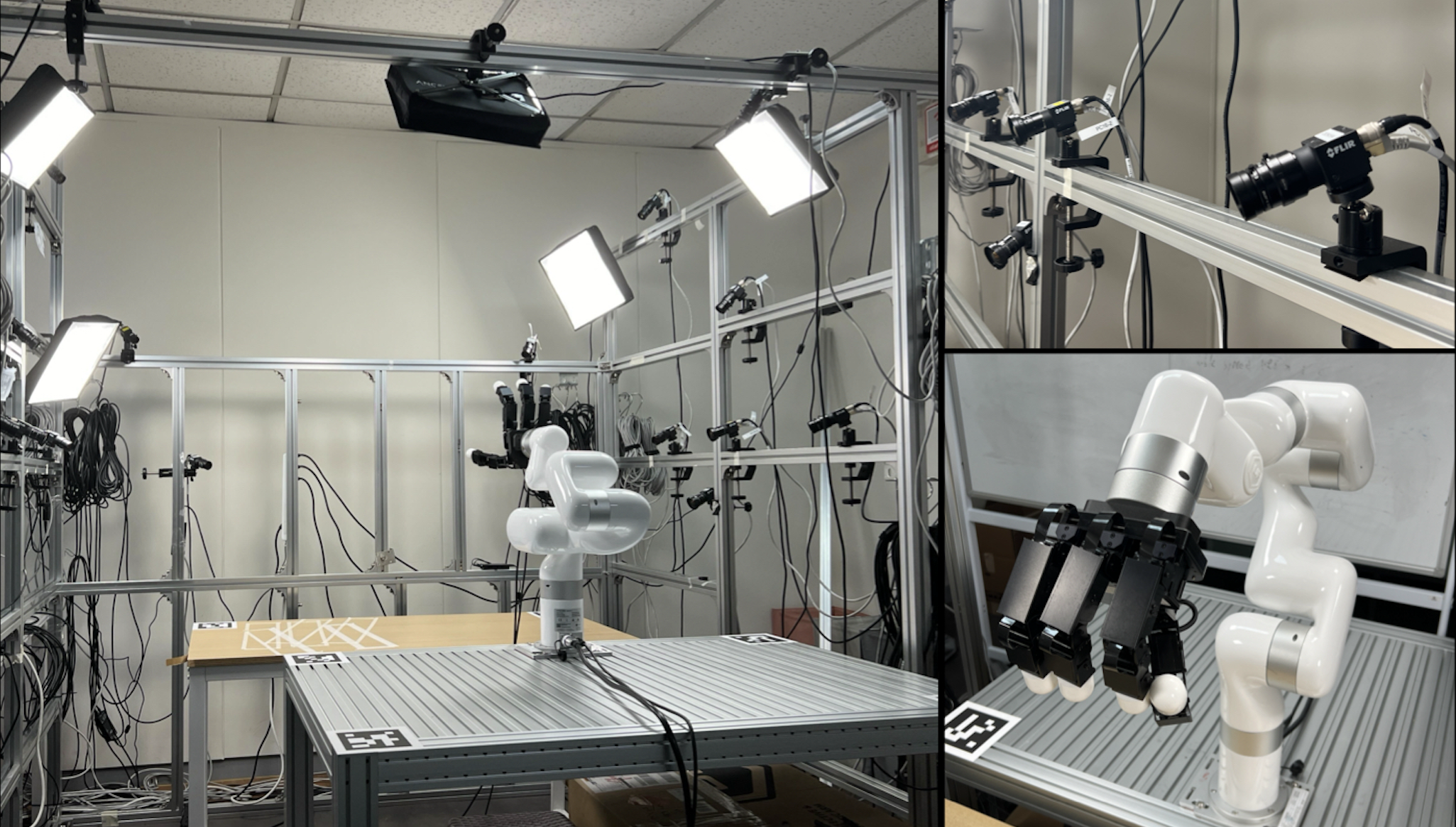}
  \caption{\textbf{System Overview:} Our system consists of 16 synchronized cameras, an xArm6 robot arm, and a 16-DoF Allegro robot hand.}
\label{fig:system}
\vspace{-10pt}
\end{figure}

\begin{figure}[t]
\centering
\includegraphics[width=0.9\columnwidth, trim={0cm 1.0cm 0cm 0.5cm}]{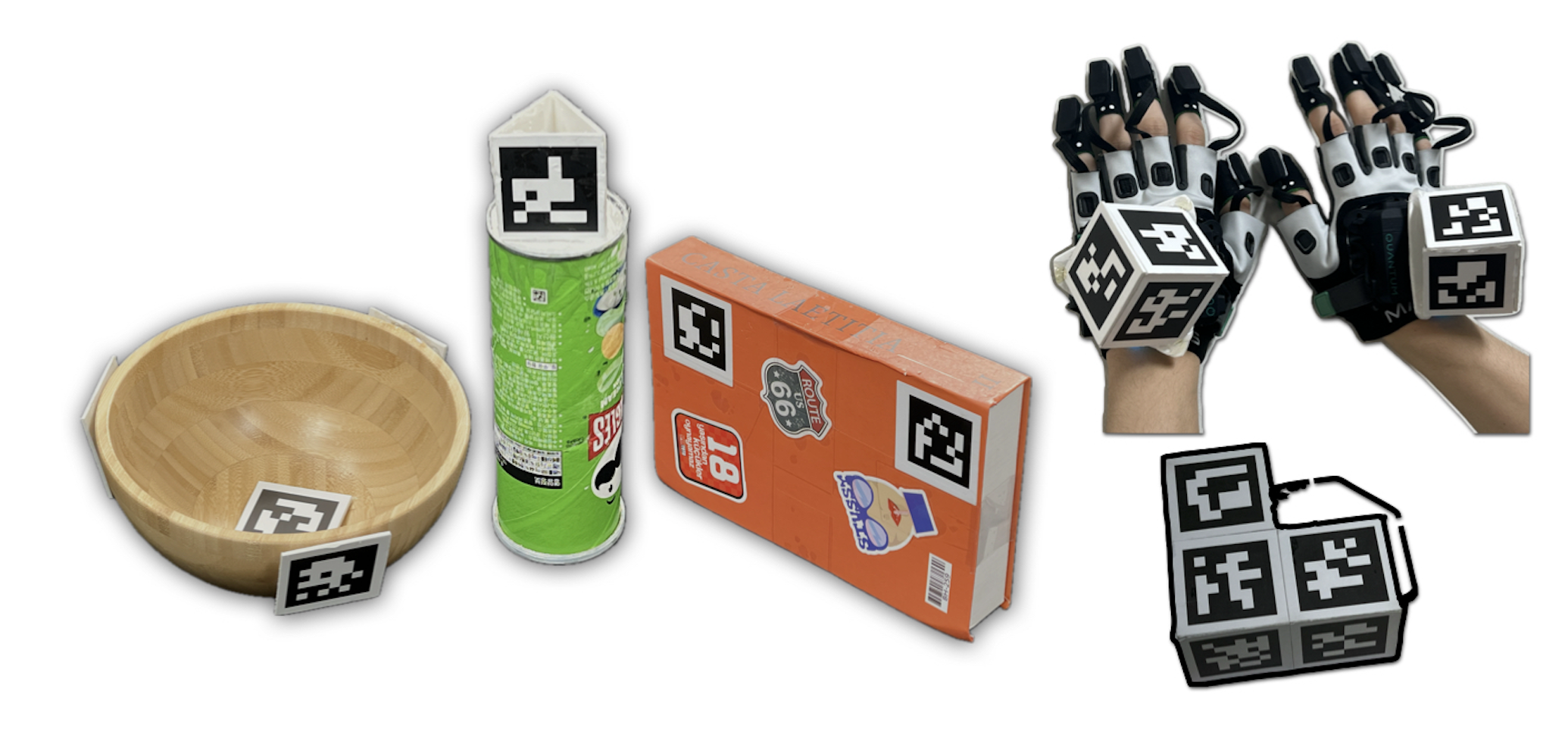 }
    \caption{Objects used in the experiment and a marker system for 3D tracking.}
\label{fig:object_task}
\vspace{-25px}
\end{figure}

To collect the required human mocap demo and robot teleoperation data, we build a multi-camera system with 16 cameras paired with wearable motion capture devices and gloves to capture 3D human body motion and object movements, as shown in Fig.~\ref{fig:system} following the system of ~\cite{kim2024parahome}. The 3D human body and hand motions cues are obtained from wearable mocap devices. The multi-camera system is used to track the 3D object movement by tracking the attached aruco markers on the object and the gloves, as shown in Fig.~\ref{fig:object_task}, where the marker on the gloves are required to align the human motion and object in the same 3D coordinate system. Multi camera system is synced and spatially calibrated. In our setup, objects, human hands, and robot arms and hands are located in the common 3D coordinate, with 30Hz capture frequency. After calibration, we collect human mocap demo and robot teleoperation data using the same system. To perform robotic teleoperation, we directly use the teleoperator's wrist pose w.r.t. pelvis frame and hand joint angles acquired from the mocap device, as a robot action. 
\vspace{-5pt}

\vspace{-5pt}
\section{Evaluations}
\label{sec:experimental_setup}

 \vspace{-5px}
\begin{figure}[b]
\centering
\includegraphics[width=0.95\columnwidth, trim={0cm 1.0cm 0cm 2.0cm}]{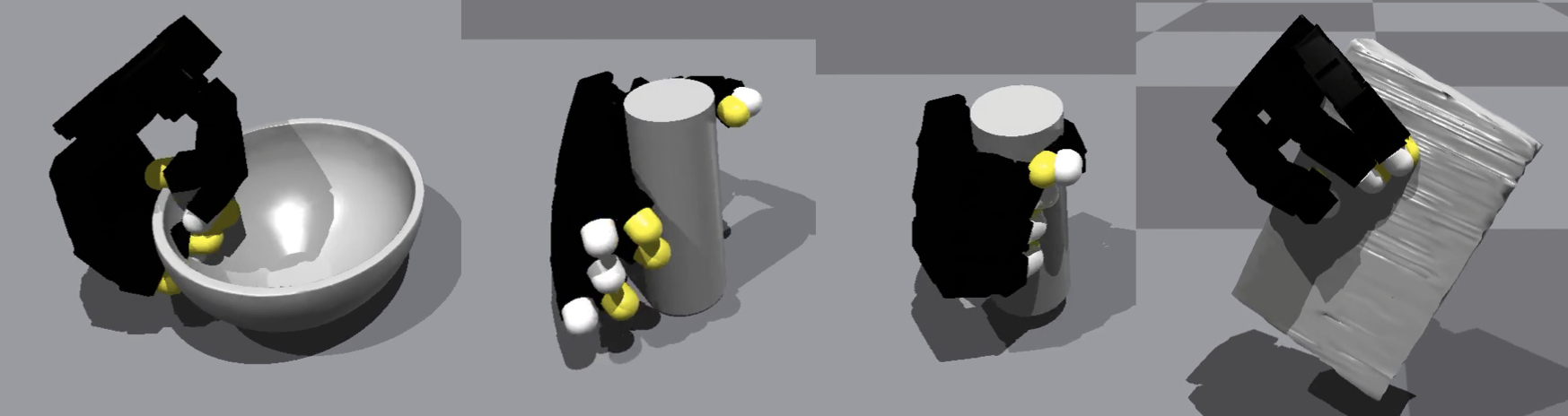}
    \caption{\textbf{Visualization of robot teleoperation dataset}. \newline
    \textbf{Yellow}: desired robot joint values. \textbf{White}: actual robot joint values. The dataset is collected in real world, and Isaac Gym simulator is only used for rendering.}
\label{fig:teleop}
\end{figure}

\subsection{Experimental Setup}
\vspace{-3pt}
We choose three objects and corresponding tasks with different characteristics to show the validity of our framework. Three objects and its corresponding tasks are as below.

\noindent \textbf{Bottle.} The robot must pick the bottle from a randomized starting location, and place it within the target location without tipping it over . The bottle's diameter is sufficiently large such that a human hand cannot fully encompass it, whereas the robot hand is capable if enveloping the whole bottle due to larger hand size. Naively matching robot and human hand fingertips may result an unstable grasp.
\textbf{Bowl.} The robot must pick the bowl from a randomized starting location while maintaining upright rotation, and place it within the target location. The bowl's concave shape induces a complicated contact interaction between human hand and object hand, and a precise control is needed for the robot in order to pick up the bowl without tilting it. \textbf{Book.} The robot must pick up the book and reorient it in order to make it stand vertically. Due to the book's flatness and size, the robot hand must maintain a fine, consistent contact with the book during reorientation to prevent slipping. Additionally, careful placement is crucial to ensure the book remains upright. 

For each task of bottle, bowl, book, we collect 113, 100, 114 human mocap demonstrations and 92, 114, 93 robot teleoperation demonstrations, respectively. There exists many different possible object trajectories achieving the same task, emphasizing the need of synthetic data generation pipeline we propose. We train $\mathbf{\Psi}_{enc,\,dec}$ and $\mathbf{\psi}_{enc,\,dec}$ per each task, and a regressor $\Omega$ is shared across tasks. Each dataset is divided into trainset and validset with ratio 9 : 1.

We use position control for controlling the robot arm and robot hand. Note that the desired wrist pose and hand joint values (i.e. robot actions that are actually fed as commands) may differ from actual values based on the interaction between robot and the object, as shown in Fig.~\ref{fig:teleop}. While the visualization of desired robot proprioception penetrates the surface, the actual robot is in contact with the object along its surface. Such discrepancy makes kinematics-based retargeting methods(i.e. matching human and robot fingertips) suboptimal, as the robot finger may slip due to inadequate contact forces.

We aim to the answer the following questions through the experiment. \textbf{(Q1)} Does a hand motion refinement process along manifold of human mocap produces more physically plausible and natural human motion for synthetic human-object interaction data? \textbf{(Q2)} Is our regression model $\Omega$  necessary to provide a good latent initialization for hand motion refinement? \textbf{(Q3)} Can our retargeting model $\mathcal{F}$ better translate human mocap data to robot action data compared to baselines? \textbf{(Q4)} Can our retargeting model $\mathcal{F}$ generalize to unseen object trajectories? \textbf{(Q5)} Is our retargeting model robust to noise in the human mocap demo?

\vspace{-5pt}
\subsection{Synthetic Paired Dataset Generation Model S }
\begin{figure}[t]
\centering
\includegraphics[width=0.9\linewidth, trim={0cm 0.3cm 0cm 0cm}]
{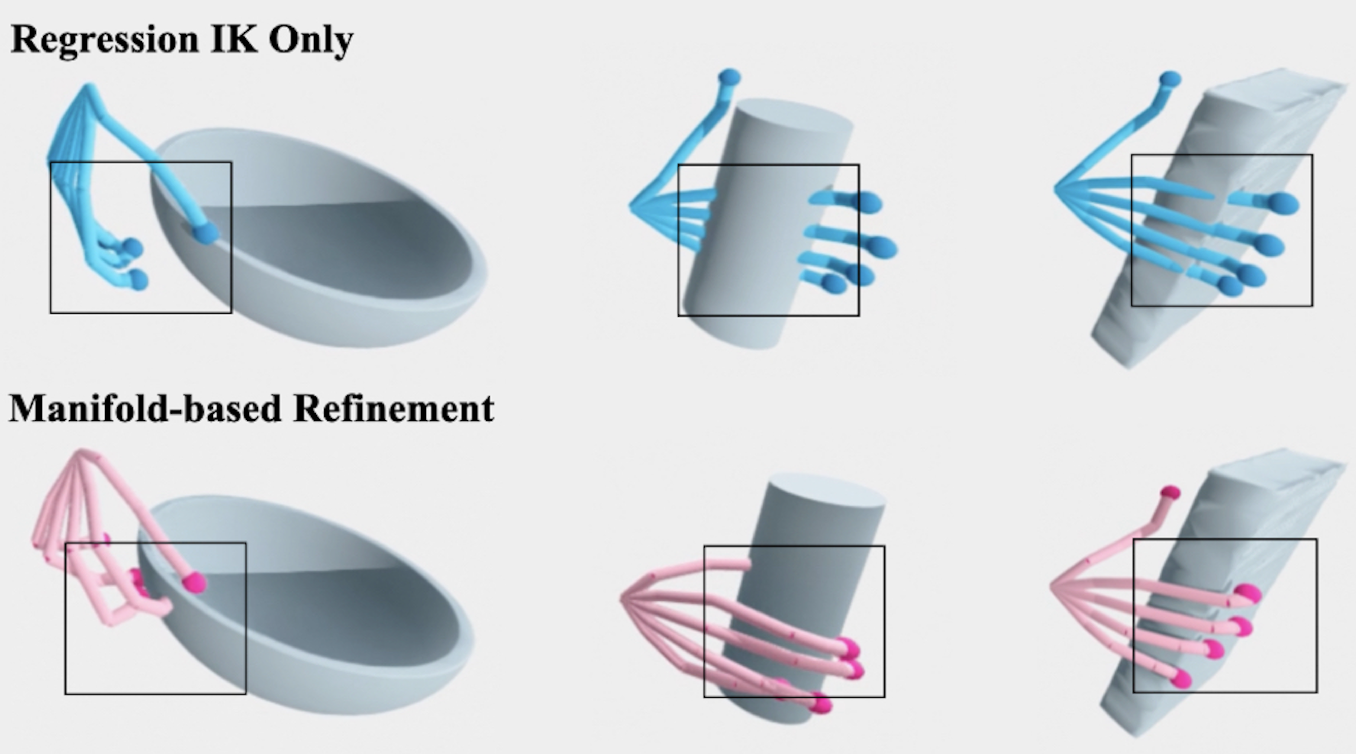}
\vspace{0px}
    \caption{Qualitative comparison before and after applying hand motion refinement model for synthetic data generation. \newline\noindent
    \textbf{Blue hand} and \textbf{red hand} indicates before and after refinement, respectively. Hand motion after refinement shows more plausible hand movements.}

\label{fig:phase1}
\vspace{-5px}
\end{figure}

\begin{table}
\centering
\footnotesize
\begin{tabular}{cccc}

\toprule[0.8pt]
Task & Method & Penetration ($m,\,\downarrow$) & Contact($m,\,\downarrow$) \\ 
& & tip / tip+mid &tip / tip+mid \\
\hline

 \multirow{2}{*}{Bottle} & Ours w/o refine. & 0.003 / \textbf{0.002} & 0.014 / 0.014 \\
 & Ours & \textbf{0.002} / \textbf{0.002} & \textbf{0.010} / \textbf{0.009} \\
\hline
 \multirow{2}{*}{Bowl} & Ours w/o refine. & \textbf{0.005} / \textbf{0.003} & 0.013 / 0.013\\ 
 & Ours & 0.009 / 0.007 & \textbf{0.010} / \textbf{0.010} \\

\hline
  \multirow{2}{*}{Book} & Ours w/o refine.& 0.012 / 0.006 & 0.037 / 0.034 \\ 
 & Ours & \textbf{0.004} / \textbf{0.004} & \textbf{0.024} / \textbf{0.026} \\
 
 \hline
  \multirow{2}{*}{Total} & Ours w/o refine. & 0.006 / \textbf{0.004} & 0.021 / 0.020\\ 
 & Ours & \textbf{0.005} / \textbf{0.004} & \textbf{0.014} / \textbf{0.014}\\
\hline

\end{tabular}
\caption{Comparison between ours and without refinement. tip indicates evaluating fingertip, and tip+mid indicates evaluating both fingertip and mid joint.
}
\label{t1}
    \vspace{-20px}
\end{table}

In this section, we verify our design choice for synthetic dataset generation pipeline.

\noindent \textbf{Effectiveness of human hand motion refinement.} First, we evaluate the performance of human hand motion refinement, based on the following metrics.

\noindent \textbf{Contact}. Measured by the distance between human fingertips or finger middle joints and closest object surface during manipulation. Middle joint refers to joint between distal and middle bone.
Within all our tasks, human's all fingertips and middle joints are naturally in contact with the target object during manipulation.

\noindent \textbf{Penetration}. Measured by the penetration depth of human fingertips or finger middle joints. A plausible human hand motion should not have penetration with the object. 

\noindent All metrics are computed on the synthetic dataset we generated, $\{O_i, H^{init}_i\}$ and $\{O_i, H_i\}$, each referring to before and applying hand motion refinement. $O_i$ is from teleoperation dataset. Metric is averaged over all frames within the duration when the object is in motion, based on the generated synthetic hand motion and the ground truth object trajectory.

Results are in Table.~\ref{t1}. Applying motion editing produces more natural and physically plausible human hand motions, reducing both contact and penetration errors in most cases. Fig.~\ref{fig:phase1} shows qualitative examples before and after applying motion editing. The motion editing model successfully corrects wrong finger positions(i.e. fingertip not contacting the bowl, fingers penetrating the book) and wrong wrist pose(i.e. human wrist rotation is inaccurate making the whole hand to penetrate the bottle). 

\begin{figure}[t]
\includegraphics[width=0.95\columnwidth, trim={0cm 0.6cm 0cm 0cm}]
{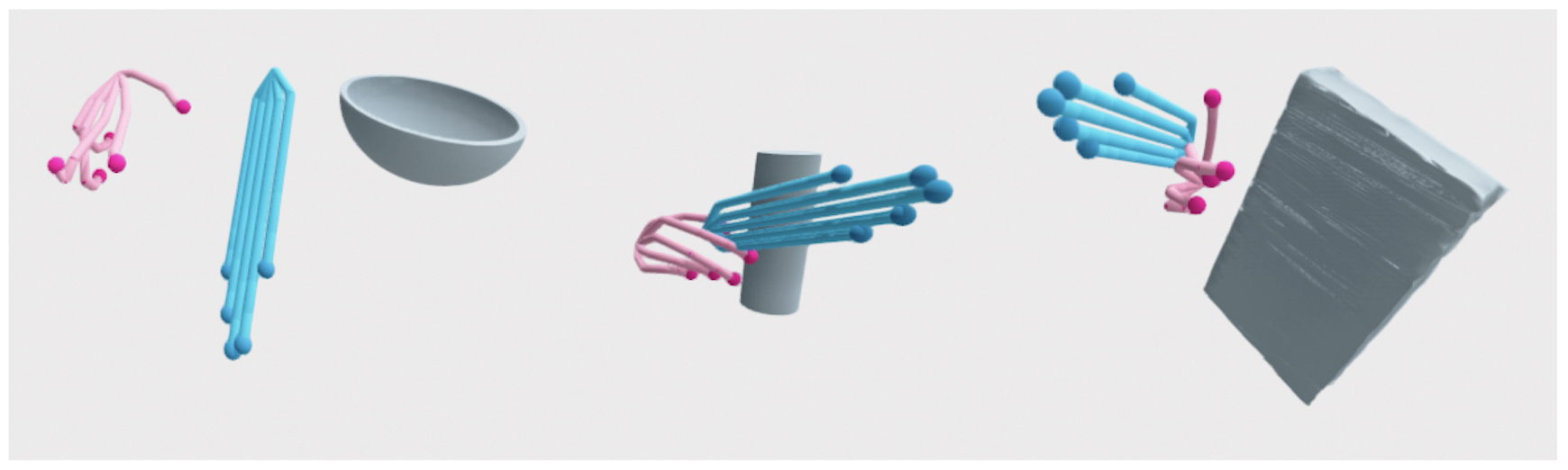}
    \caption{Ablation on different initial estimate before refinement. \textbf{Blue hand} is the initial estimate, which uses robot wrist pose and zero-pose human fingers instead of our regression model. \textbf{Red hand} is the refined hand motion.}
\label{fig:phase1_ablation}
\vspace{-10px}
\end{figure}

\noindent \textbf{Necessity of regression-based IK model.} To 
demonstrate the effectiveness of initializing based on regression-based IK model, we compare an alternative initialization which uses robot wrist pose and zero-pose human fingertip positions 
w.r.t. human wrist frame. 
As shown in Fig.~\ref{fig:phase1_ablation}, as zero hand initial pose are far away from plausible hand motion, latent optimization becomes unstable and generates unnatural hand and object interaction.

\vspace{-6pt}
\subsection{Human-to-Robot Retargeting Model F }
\vspace{-4pt}
In this section, we evaluate our Retargeting Model $\mathcal{F}$ within the aforementioned three tasks, with the following metrics.

\begin{table}
\centering
\footnotesize
\begin{tabular}{cccc}

\toprule[0.8pt]
Task & Method & COM Error & Ori Error  \\
& & ($m,\,\downarrow$) & ($rad,\,\downarrow$) \\
\hline

 \multirow{3}{*}{Bottle} & Fingertip & 0.097 & 0.31 \\
 & Fingertip + Midjoint & 0.085 & \textbf{0.15} \\
 & Ours & \textbf{0.050} & 0.22 \\
\hline
 \multirow{3}{*}{Bowl} & Fingertip & \textbf{0.051}(\textbf{0.052}) & 0.68(0.72) \\ 
 & Fingertip + Midjoint & 0.058 & 0.72 \\
 & Ours & 0.054(0.054) & \textbf{0.57}(\textbf{0.49}) \\
\hline
  \multirow{3}{*}{Book} & Fingertip & 0.055 & 0.24 \\ 
 & Fingertip + Midjoint & 0.062 & 0.25 \\
 & Ours & \textbf{0.052} & \textbf{0.22} \\
 \hline
  \multirow{3}{*}{Total} & Fingertip & 0.067 & 0.42 \\ 
 & Fingertip + Midjoint & 0.068 & 0.39 \\
 & Ours & \textbf{0.052} & \textbf{0.34} \\

\hline
\end{tabular}
\caption{Error metrics for our method and baselines. 
For the Bowl task, we also compare the robustness of our method with the Fingertip baseline with noisy input. Values in parentheses indicate when the human hand motion is from noised mocap.
}
\vspace{-3pt}
\label{t2}
    \vspace{-20px}
\end{table}

\noindent \textbf{COM and Ori Error}: These two terms measure the $L2$ norm between the object's target trajectory and the measured trajectory using our retargeting model, as in \cite{dasari2023learning}. We compute the error within succeeded trajectories, as failure trajectories may show random object motion or no object motion, which makes error term less relevant to the model's performance (e.g. it is hard to compare object not moving at all vs. object being tipped over during grasping.).

\noindent \textbf{Success Rate and Number of Completed Subtasks}. For each task, we define task-specific metric and success criteria. The sub tasks are divided into pick and place, where 'pick' is determined by whether the robot successfully lifts the object(with all points off the ground), and 'place' is determined by whether the robot successfully places the object while maintaining balance. We evaluate as success when both the pick and place tasks were completed. For bowl task, we have an additional constraint to rotation: if the bowl tilts more than 45 degrees in the direction of the object's z-axis, it is considered a failure.

\noindent \textbf{Baselines.} We consider two types of baselines which are commonly used when learning robot policy from human mocap data or when performing robotic teleoperation. \textbf{Fingertip Matching}, which uses optimization-based IK to directly match robot hand fingertips to human hand fingertips, and \textbf{Fingertip and Middle Joint Matching}, which has additional constraint which enforces middle joints of robot hand and human hand to be in an identical position. Baselines do not take the interacting object into account, but rather considers the human hand's kinematic information only.

\noindent \textbf{Generalization capability.} To check whether our retargeting model generalizes to unseen human hand motion and object trajectory, we do the following evaluation. First, we split the human mocap data $\{O,H\}$ into train dataset and validation dataset, and use trainset only for human hand refinement model training. Then, the validset is used for evaluating the retargeting model $\mathcal{F}$. Note that we do not assume any paired ground truth robot action for both trainset and validset(i.e. both sets can be unseen to our retargeting model), but rather utilize the synthetic data we generated.

Table.~\ref{t2} and Table~\ref{t3} show the evaluated results of our model and baselines. Each task was evaluated over 10 episodes for the Bottle and Bowl task, and 11 episodes for the Book task. Overall, our method consistently achieved lower COM and Ori errors, resulting in a significantly lower total loss compared to the baselines. Our model also achieved higher success rate and sub task completion. While the baseline methods occasionally outperformed ours in either accuracy or task success, there was no instance where they demonstrated superior performance in both aspects simultaneously. Considering both aspects together and regarding overall robustness, our method outperforms. 

\textbf{Robustness to noisy mocap demo.} While our system utilizes a multi-view camera system for accurate mocap, it is not always a viable option for collecting human mocap data. A promising alternative is to use vision-based human-objects reconstruction models. However, these models are often inaccurate, producing noisy human hand and finger poses. As we aim to develop a framework that can scale robotic data by retargeting human motions, we evaluate our model's robustness with noisy mocap data. Specifically, we choose Bowl task and Fingertip as baseline to compare against our method, as it has best performance among baselines and tasks. We add a gaussian noise to the validset we used for evaluation above with mean=0.001 at all dimension of human hand, from wrist pose to finger positions. The results are in Table.~\ref{t2} and Table.~\ref{t3}, where the values in parentheses indicate evaluation under noisy mocap. Surprisingly, although we add a very small noise, the performance of Fingertip baseline drops from 0.5 to 0.3 in terms of success rate, while ours got even better, from 0.8 to 0.9 success rate. COM and Ori Error stays similar, as it is computed over succeeded trajectories. This intuitively shows the instability of kinematics based baseline, which assumes mocap to be accurate. As our model learns the manifold of human hand and object interaction itself(i.e., the model implicitly knows robot hand or human hand should be in contact with the object when it is moving.), it shows robustness to certain level of noise.

\begin{table}
\centering
\footnotesize
\begin{tabular}{cccc}
\toprule[0.8pt]
Task & Method & \#  Subtasks & Success \\
\midrule

 \multirow{3}{*}{Bottle} & Fingertip & 1.0 & 0.2 \\
 & Fingertip + Midjoint & 0.6 & 0.2 \\
 & Ours & \textbf{1.7} & \textbf{0.7} \\
\hline
 \multirow{3}{*}{Bowl} & Fingertip & 1.4(1.1) & 0.5(0.3) \\ 
 & Fingertip + Midjoint & 1.0 & 0.2 \\
 & Ours & \textbf{1.8}(\textbf{1.9}) & \textbf{0.8}(\textbf{0.9}) \\
\hline
  \multirow{3}{*}{Book} & Fingertip & \textbf{1.36} & \textbf{0.45} \\ 
 & Fingertip + Midjoint & 1.36 & 0.36 \\
 & Ours & 1.08 & 0.27 \\
 \hline
  \multirow{3}{*}{Total} & Fingertip & 1.26 & 0.39 \\  
 & Fingertip + Midjoint & 0.99 & 0.25 \\
 & Ours & \textbf{1.54} & \textbf{0.59} \\
\hline
\end{tabular}
\caption{Number of completed subtasks and success rate for our method and baselines. Values in parentheses indicate when the human hand motion is from noised mocap.}
\label{t3}
    \vspace{-23px}
\end{table}

\vspace{-7pt}
\section{Discussion and Limitations}
\label{sec:conclusion}
\vspace{-5px}
In this work, we developed a framework for learning a retargeting model which translates human mocap demo to a sequence of plausible robot actions for reproducing the manipulation. Under a carefully designed pipeline, we achieve superior performance to baselines in multiple real world dexterous manipulation tasks, even within noisy mocap data. While our framework showed generalization capabilities along different trajectories within the same object, we have separate models for each object. Moreover, we only use object pose to represent each object. Learning a general, unified retargeting model along with rich representation induced from mocap data (i.e., proximity between hand and object) with a more scaled experiment will be an interesting future direction to pursue.

\vspace{-6px}
\section*{ACKNOWLEDGMENT} 
\vspace{-5px}
This work was supported by NRF grant funded by the Korean government (MSIT) (No.
2022R1A2C2092724, No. RS-2023-00218601), and IITP grant funded by the Korea government (MSIT) [No. RS-2024-00439854 , No. RS-2021-II211343]. H. Joo is the corresponding author.

\bibliographystyle{unsrt}
\bibliography{bib/reference}

\end{document}